\documentclass[times, review, 10pt]{elsarticle}

\usepackage{multirow}
\usepackage{amssymb}
\usepackage{amsmath}
\usepackage{bm}
\usepackage[dvipsnames]{xcolor} 
\usepackage[hidelinks]{hyperref}

\makeatletter

\usepackage{xcolor}
\usepackage{etoolbox}

\newcommand*\bluebibkeys{} 
\listadd\bluebibkeys{hu2025federated}
\listadd\bluebibkeys{unet}
\listadd\bluebibkeys{nnunet}
\listadd\bluebibkeys{unet++}
\listadd\bluebibkeys{xie2021cotr}
\listadd\bluebibkeys{butoi2023universeg}
\listadd\bluebibkeys{song2024spgnet}
\listadd\bluebibkeys{zhang2022busis}
\listadd\bluebibkeys{li2025ckdf}
\listadd\bluebibkeys{hu2025dynamic}
\listadd\bluebibkeys{li2025enhance}
\listadd\bluebibkeys{DBLP:conf/sipaim/WangHXY23}
\listadd\bluebibkeys{wu2024medsegdiff}
\listadd\bluebibkeys{wu2019consensus}
\listadd\bluebibkeys{kingma2014adam}
\listadd\bluebibkeys{}
\listadd\bluebibkeys{}
\listadd\bluebibkeys{}

\makeatletter
\newif\if@bluebibopen
\@bluebibopenfalse
\newcommand{\bluebib@close}{%
  \if@bluebibopen\endgroup\global\@bluebibopenfalse\fi
}

\let\bluebib@oldbibitem\bibitem
\renewcommand{\bibitem}[2][]{%
  \bluebib@close
  \ifinlist{#2}{\bluebibkeys}{\begingroup\color{blue}\global\@bluebibopentrue}{}%
  \bluebib@oldbibitem[#1]{#2}%
}
\AtEndEnvironment{thebibliography}{\bluebib@close}
\makeatother
\begin{document}

\begin{frontmatter}



\title{Structure and Progress Aware Diffusion for Medical Image Segmentation}
\author[1,3,4,5]{Siyuan Song}
\author[1,3,4,5]{Guyue Hu\corref{cor1}}
\author[1,3,4,5]{Chenglong Li\corref{cor1}}
\cortext[cor1]{Corresponding authors: \\
\phantom{xxx} guyue.hu@ahu.edu.cn, lcl1314@foxmail.com}
\author[1,4,5]{Dengdi Sun}
\author[1,4]{Zhe Jin}
\author[2,5]{Jin Tang}

\affiliation[1]{organization={School of Artificial Intelligence},
                addressline={Anhui University}, 
                postcode={230601}, 
                city={Hefei},
                country={China}}
                
\affiliation[2]{organization={School of Computer Science and Technology},
                addressline={Anhui University}, 
                postcode={230601}, 
                city={Hefei},
                country={China}}

\affiliation[3]{organization={State Key Laboratory of Opto-Electronic Information Acquisition and Protection Technology},
                addressline={Anhui University}, 
                city={Hefei},
                postcode={230601}, 
                country={China}}


\affiliation[4]{organization={Anhui Provincial Key Laboratory of Security Artificial Intelligence},
                addressline={Anhui University}, 
                city={Hefei},
                postcode={230601}, 
                country={China}}

\affiliation[5]{organization={Anhui Provincial Key Laboratory of Multimodal Cognitive Computation}, 
                addressline={Anhui University}, 
                postcode={230601}, 
                city={Hefei},
                country={China}}


\begin{abstract}
Medical image segmentation is crucial for computer-aided diagnosis, which necessitates understanding both coarse morphological and semantic structures, as well as carving fine boundaries. The morphological and semantic structures in medical images are beneficial and stable clues for target understanding. While the fine boundaries of medical targets (like tumors and lesions) are usually ambiguous and noisy since lesion overlap, annotation uncertainty, and so on, making it not reliable to serve as early supervision. However, existing methods simultaneously learn coarse structures and fine boundaries throughout the training process. In this paper, we propose a structure and progress-aware diffusion (SPAD) for medical image segmentation, which consists of a semantic-concentrated diffusion (ScD) and a boundary-centralized diffusion (BcD) modulated by a progress-aware scheduler (PaS). Specifically, the semantic-concentrated diffusion introduces anchor-preserved target perturbation, which perturbs pixels within a medical target but preserves unaltered areas as semantic anchors, encouraging the model to infer noisy target areas from the surrounding semantic context. The boundary-centralized diffusion introduces progress-aware boundary noise, which blurs unreliable and ambiguous boundaries, thus compelling the model to focus on coarse but stable anatomical morphology and global semantics. Furthermore, the progress-aware scheduler gradually modulates noise intensity of the ScD and BcD forming a coarse-to-fine diffusion paradigm, which encourage focusing on coarse morphological and semantic structures during early target understanding stages and gradually shifting to fine target boundaries during later contour adjusting stages. Eventually, the proposed SPAD achieves state-of-the-art performance on two medical segmentation benchmarks (AMD-SD and CXRS datasets), demonstrating the effectiveness and superiority of our approach.
\end{abstract}



\begin{keyword}
Medical image segmentation, diffusion model, structure-aware, progress-aware

\end{keyword}

\end{frontmatter}




\section{Introduction}
Medical image segmentation is a fundamental task in computer-aided diagnosis, playing a vital role in various clinical applications\cite{hu2025federated}, such as epicardial fat segmentation~\cite{liu2021anatomy} and brain tumor delineation~\cite{sun2021segmentation}. Medical segmentation not only facilitates more precise diagnoses but also alleviates the workload of healthcare professionals. In recent years, deep learning has greatly advanced this field, with U-Net~\cite{unet} and its variants~\cite{nnunet,unet++} becoming the cornerstone network architectures for many segmentation tasks.\cite{wang2026pf2smis,zhao2026active} After that, transformer-based models~\cite{xie2021cotr,dalmaz2022resvit} and large medical foundation models~\cite{wu2025medical,butoi2023universeg,r2} have further boosted the segmentation performance across various medical modalities.

\begin{figure*}
    \centering
    \includegraphics[width=0.95\linewidth]{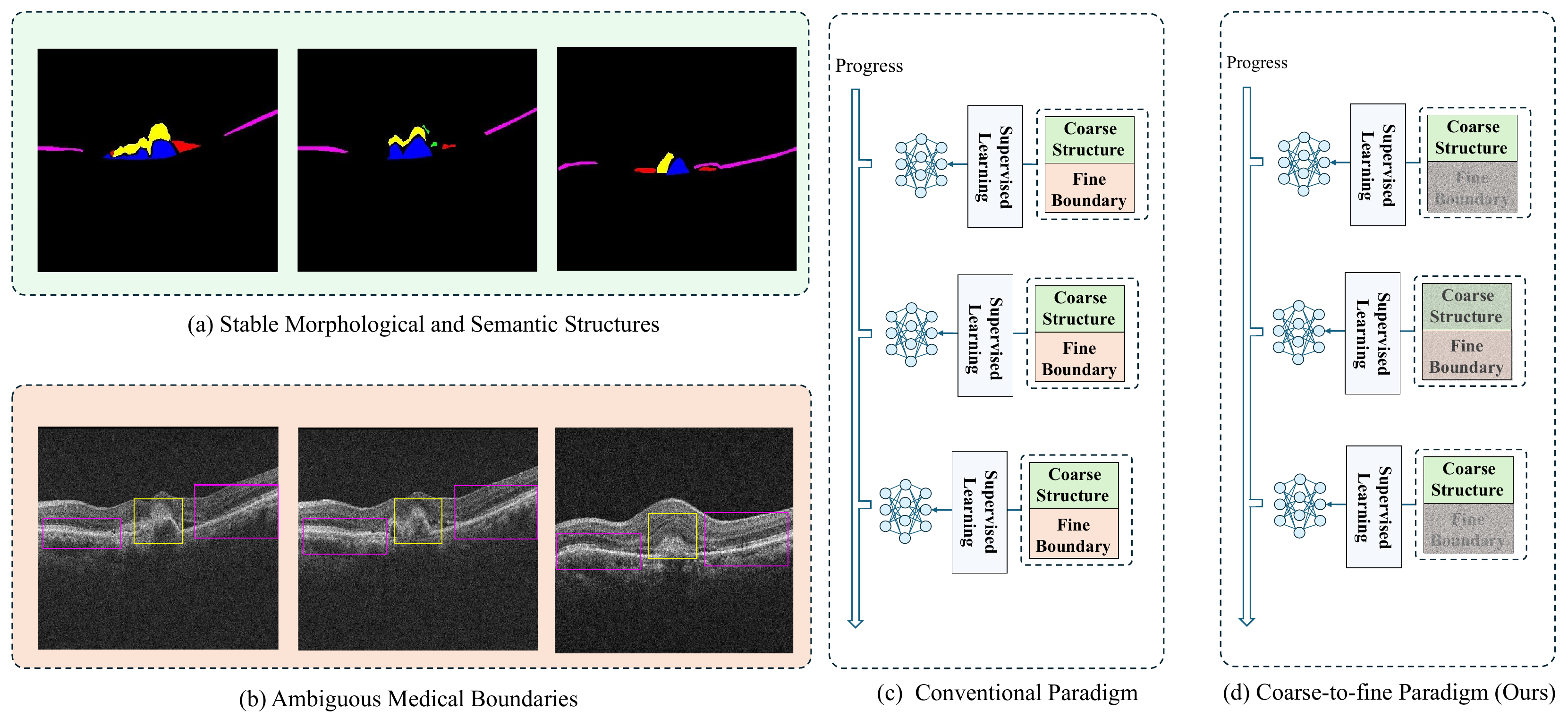}
    \caption{
    (a) Beneficial and stable morphological and semantic structures. (b) Ambiguous and unreliable medical boundaries. (c) Conventional training paradigm simultaneously learns coarse structures and fine boundaries throughout the whole training process. (d) The proposed coarse-to-fine paradigm prioritizes acquiring coarse and stable structures and gradually shifts to carving fine and unreliable boundaries in medical images.
    }
    \label{figure1}
\end{figure*}

Along with these huge advancements in network architectures \cite{unet,nnunet}, perceiving and utilizing the intrinsic structural information in medical images has also attracted increasing attention \cite{azad2024beyond,hu2025dynamic}. In detail, the structural information in medical images can be broadly categorized into morphological structure and semantic structure. The morphological structure of medical organs or lesions (such as shapes and sizes) provides essential clues for characterizing the targets and distinguishing them from surrounding tissues. Meanwhile, the semantic structure of medical organs or lesions (such as semantic classes, spatial layouts, and relative positions to neighboring anatomical organs) also offers abundant contextual guidance. For example, as in Fig.~\ref{figure1}(a), certain categories of lesions usually appear in characteristic regions and morphology, and also maintain stable positional dependencies with adjacent tissues, which contain abundant morphological and semantic structures that are beneficial for medical segmentation. Some pioneer methods have attempted to exploit this structural information in medical images to enhance structural awareness during segmentation. On one hand, some approaches explicitly incorporate structural priors or encode spatial relationships to strengthen the understanding of global context and inter-class consistency in segmentation models ~\cite{cheng2022contour,song2024spgnet,hu2025dynamic,rachmadi2024improving}. On the other hand, some methods also implicitly capture structural dependencies via architectural innovations or attention mechanisms to better recognize spatial patterns and contextual relations~\cite{azad2024beyond}. Together, these structure-aware strategies highlight the benefits of leveraging morphological and semantic structures for precise medical image segmentation.

Medical image segmentation requires not only understanding coarse morphological and semantic structures but also carving fine boundaries. 
\textit{Unlike the morphological and semantic structures that are always beneficial and stable for target understanding}, 
\textit{the fine boundaries of medical targets (such as tumors and lesions) are usually ambiguous and noisy}. 
As illustrated in Fig.~\ref{figure1}(b), the fine boundaries of tumors and lesions are frequently blurred and ambiguous for various reasons, such as lesion overlap~\cite{mendes2023avoiding}, annotation uncertainty~\cite{armato2011lung}, low contrast~\cite{zhang2022busis}, etc. 
Due to boundary ambiguity, the fine boundary information of medical targets is noisy and not always reliable as trustworthy supervision, particularly in the early target understanding stage. 
However, existing methods~\cite{xing2023diff,nnunet} that simultaneously learn coarse structures and fine boundaries throughout the whole training process (as shown in Fig.~\ref{figure1}(c)) may be sub-optimal under such boundary uncertainty. 
Therefore, a more reasonable learning paradigm to coordinate coarse structures and fine boundaries should prioritize the acquisition of coarse and stable structures during early target understanding stages, and gradually shift to fine and unreliable boundary adjustment during the later contour carving stages (as shown in Fig.~\ref{figure1}(d)), a stage-aware strategy that has also been explored in other learning frameworks to improve robustness and representation stability across training stages~\cite{li2025ckdf,li2025enhance,hu2023compositional}.

To address these issues, we propose a structure and progress aware diffusion model (SPAD) for medical image segmentation. It consists of two key components of semantic-concentrated diffusion (ScD) and boundary-centralized diffusion (BcD) modulated by a progress-aware scheduler (PaS). The ScD introduces anchor-preserved target perturbation by perturbing pixels within a specific medical target but preserving small unaltered areas as semantic anchors, guiding the model to infer the corrupted regions from the surrounding semantic context. The BcD introduces progress-aware noise to the boundary regions, suppressing the influence of unreliable or ambiguous boundaries of medical targets, encouraging the model to focus on learning coarse anatomical morphology and global semantics during early stages. The progress-aware scheduler (PaS) gradually modulates noise intensity in both ScD and BcD throughout the training process, ensuring a smooth transition from understanding coarse morphological and semantic structures to carving fine boundaries.

In summary, the main contributions of this paper are 4 folds:

\begin{itemize}
  \item We propose a novel structure and progress aware diffusion  (SPAD) for medical image segmentation, which makes use of morphological and semantic structures and alleviates the boundary ambiguity as well.
  \item We introduce a semantic-concentrated diffusion (ScD) mechanism, which guides inferring corrupted areas from the surrounding semantic context and improves the inter-target structural reasoning and anatomical rationality.
  \item We propose a boundary-centralized diffusion (BcD) mechanism, which suppresses unreliable medical boundaries and focuses on learning coarse anatomical morphology and global semantics during early stages.
  \item We propose a progress-aware scheduler (PaS) tailored for diffusion-based segmentation models, which encourages a coarse-to-fine paradigm that prioritizes acquiring coarse and stable structures and gradually shifts to carving fine and unreliable boundaries.

\end{itemize}

\section{Related Works}

\subsection{Structure-Aware Segmentation}
Precision medical image segmentation requires capturing both fine-grained object boundaries and high-level semantic structures, such as global shapes and inter-target spatial relationships. Boundary information is often ambiguous due to low contrast, noise, or annotation uncertainty, while structural cues are generally more stable and informative, making effective utilization of both types of information crucial for robust segmentation.
Existing approaches address these challenges in two main ways. One line of work introduces explicit structural priors to guide the model: SPGNet \cite{song2024spgnet} employs PCA-based statistical shape models, DSAIF \cite{gu2025dual} applies Max-tree/Min-tree topological filtering, DSC‑AMP \cite{hu2025dynamic} dynamically adapts convolutional kernels to anatomical morphology. SPM \cite{DBLP:journals/tmi/YouHYG25} injects global and local shape priors in a U-Net framework, and a recent shape-intensity knowledge distillation method \cite{DBLP:journals/fcsc/DongDX25} transfers shape-intensity knowledge via teacher-student training for improved cross-domain robustness. Another line of work encodes structural information implicitly through architectural design: D‑LKA Net\cite{azad2024beyond} captures local and global context with deformable large-kernel attention, Swin‑DeformAttn \cite{DBLP:conf/sipaim/WangHXY23} and DeformUX‑Net \cite{lee2023deformux} leverage deformable and large-kernel operators to enhance anatomical consistency without extra supervision, and multi-scale fine-grained feature extraction has also been explored \cite{zhang2019scalenet}. Style or domain adaptation methods also maintain structural consistency across domains but do not provide direct guidance during training.
Despite advancements, most existing methods supervise both structure and boundary regions simultaneously, which can lead to suboptimal learning when boundaries are ambiguous. In contrast, our approach decouples structure learning from boundary refinement via a progress-aware training schedule. This enhances robustness and accuracy, especially in scenarios with uncertain or ambiguous boundaries.

\subsection{Diffusion Models for Medical Image Analysis}

Diffusion models have recently gained prominence in medical image analysis due to their ability to model complex data distributions, handle uncertainty, and generate high-fidelity anatomical structures. These properties make them suitable for a wide range of tasks, including image reconstruction, denoising, data augmentation, and segmentation \cite{kazerouni2023diffusion,webber2024diffusion,pan2025diverse}. A key challenge across these applications is ensuring that the generated or reconstructed images preserve both global anatomical structures and fine-grained details, particularly in regions with low contrast, noise, or ambiguous boundaries. Diffusion models have shown significant potential in overcoming these challenges, thereby advancing the state-of-the-art in medical image processing.
Existing works can be broadly categorized into two main directions \cite{kazerouni2023diffusion,shi2024diffusion}. The first direction focuses on task-driven applications \cite{wu2024medsegdiff,wu2024medsegdiff2,xing2023diff}, where diffusion models are applied to specific medical image tasks, such as segmentation, lesion detection, and organ localization. For example, MedSegDiff~\cite{wu2024medsegdiff} and MedSegDiff-V2~\cite{wu2024medsegdiff2} incorporate dynamic conditional encoding and transformer-based components, improving the model’s ability to handle complex anatomical variations and provide accurate segmentations. Similarly, Diff-UNet~\cite{xing2023diff} introduces step-uncertainty fusion in a 3D U-Net to improve robustness in volumetric prediction, which is particularly useful for tasks that involve three-dimensional medical data. These existing diffusion-based segmentation methods \cite{wu2024medsegdiff,wu2024medsegdiff2} mainly improve performance through architectural conditioning or inference-stage aggregation, and they generally apply uniform noise modeling during training. In contrast, our approach focuses on regulating the training process itself by introducing region-aware perturbations and a progress-guided learning strategy, explicitly decoupling structural learning from boundary refinement in a coarse-to-fine manner. These models are designed to optimize task performance and adapt to the specific challenges presented by different medical imaging applications. The second direction focuses on training process and strategy optimization \cite{wu2019consensus,mayo2024denoising}, where the emphasis is on enhancing the model's learning capabilities through novel training strategies. The methods in this category often introduce mechanisms such as boundary-region noise injection or selective perturbations to guide the learning process. For instance, noise-to-noise frameworks like SSIM~\cite{wu2019consensus} and other denoising methods have been explored to refine image generation and restoration. These strategies enhance the model's robustness towards noise and enable it to generate high-quality images from noisy data. Additionally, some methods (such as MRF-IDDPM~\cite{mayo2024denoising}) focus on improving image reconstruction and enhancing structure consistency by modulating the noise injection process. These techniques optimize the way the model learns from noisy and ambiguous data, ensuring better generalization across diverse medical imaging tasks. Our method uses diffusion models to separate structure and boundary learning, improving segmentation accuracy. By prioritizing structure learning and boundary refining successively, it performs well in challenging cases with ambiguous boundaries or complex structures.

\begin{figure*}
    \centering
    \includegraphics[width=\linewidth]{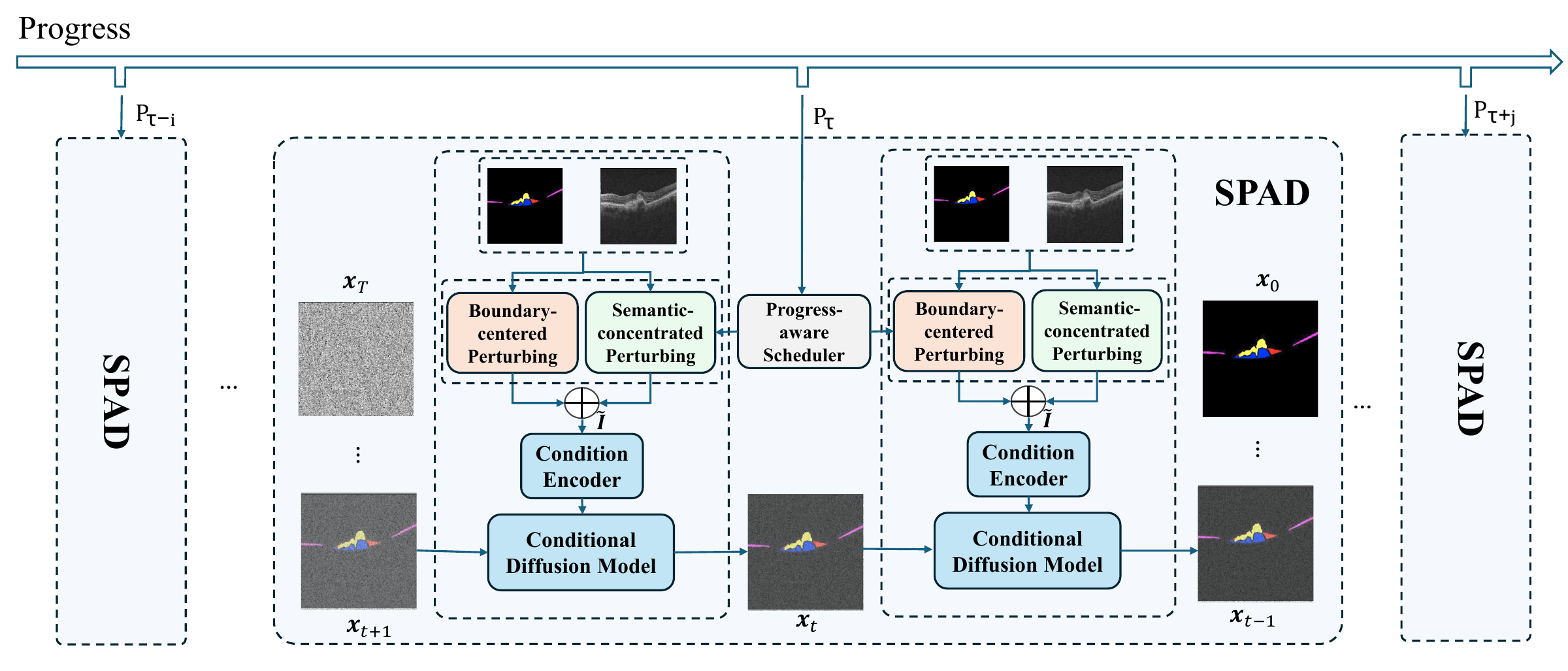}
    \caption{
Overview of the proposed structure and progress-aware diffusion (SPAD) framework. The conditional diffusion model progressively denoises the final segmentation prediction $\textbf{x}_0$ from a random noise $\textbf{x}_T$ (initial noise input) and the perturbed segmentation image (progress-aware noise condition) step by step. The progress-aware noise condition is generated from the combination of boundary-centered perturbing and semantic-concentrated perturbing. At each step, the perturbing intensities are dynamically modulated by a progress-aware scheduler (PaS), enabling a progress-aware coordination of the coarse structures and fine boundaries in medical segmentation targets.}
    \label{framework}
\end{figure*}

\section{Methods}
\subsection{Preliminary}

Diffusion models comprise a forward noising process and a reverse denoising process. During the forward process, Gaussian noise is progressively injected into the ground-truth segmentation label \(\textbf{x}_0\), producing a sequence of increasingly corrupted representations \(\{\textbf{x}_1, \textbf{x}_2, \ldots, \textbf{x}_T\}\) over \(T\) discrete timesteps. It defines a Markov chain where each state depends only on its predecessor. In contrast, the reverse process aims to learn the conditional distributions that recover the clean label map from the noisy observations by reversing this degradation step-by-step. The generative process factorizes as:

\begin{equation}
p(\textbf{x}_{0:T-1} \mid \textbf{x}_T) = \prod_{t=1}^{T} p(\textbf{x}_{t-1} \mid \textbf{x}_t),
\end{equation}
where each term \(p(\textbf{x}_{t-1} \mid \textbf{x}_t)\) is parameterized by a neural network trained to approximate the inverse of the forward noising step. The reverse process begins with a sample drawn from an isotropic Gaussian distribution, \(p(\textbf{x}_T) = \mathcal{N}(0, \mathbf{E}_{n \times n})\), where \(\mathbf{E}\) denotes the identity matrix. Over time, this sample is refined through iterative denoising steps, guided by the learned model \(p\), until it converges to a segmentation map.

To adapt diffusion models for semantic segmentation, the denoising process is conditioned on an input image $\bm{I}$. This conditioning allows the model to incorporate structure cues from $\bm{I}$ while recovering the segmentation mask. Specifically, a noise prediction function \(\epsilon(\textbf{x}_t, \bm{I}, t)\) is learned, which estimates the noise added at timestep \(t\), given the noisy segmentation state \(\textbf{x}_t\), the image \(\bm{I}\), and the timestep index \(t\). The condition image and the current segmentation estimate are first embedded into latent features, which are fused to predict the noise:

\begin{equation}
\epsilon(x_t, \bm{I}, t) = f(\phi(x_t), \psi(\bm{{I}}), t),
\end{equation}
where \(\phi(\cdot)\) and \(\psi(\cdot)\) denote feature extractors for the noisy label and the input image respectively, and \(f_{\theta}(\cdot)\) is a denoising function that operates on these features together with the embedded timestep \(t\). The final segmentation map is obtained by repeating the denoising process from \(t = T\) to \(t = 0\).

\subsection{Overview of Structure and Progress Aware Diffusion}

The overview pipeline of the proposed structure and progress aware diffusion (SPAD) framework for medical image segmentation is shown in Fig.~\ref{framework}. The SPAD is built upon a conditional diffusion backbone that generates segmentation maps conditioned on the medical image input. During training, the input image and its ground-truth mask are routed in parallel to two perturbation modules: semantic-concentrated perturbing (ScP) and boundary-centered perturbing (BcP). Semantic-concentrated diffusion (ScD) and boundary-centralized diffusion (BcD) are primarily realized by ScP and BcP, respectively, while their behaviors are regulated by the Progress-aware Scheduler (PaS).

In detail, the ScD strategy perturbs selected target regions while preserving anchor areas. This design enables the model to exploit semantic context from surrounding targets while retaining stable reference signals, thereby enhancing global structural reasoning and improving lesion localization. In contrast, the BcD strategy focuses on contour regions by injecting noise into uncertain boundaries. It guides the model to pay more attention to ambiguous edges and progressively refine fine-grained details, resulting in sharper and more coherent boundary delineations. Both the ScD and BcD are coordinated by the progress-aware scheduler, which modulates perturbation strength across training epochs by gradually decreasing the perturbation intensity to enable a transition from learning coarse anatomical structures to refining fine boundary details.

Overall, ScD and BcD introduce two different uncertainties in medical segmentation (semantic incompleteness and boundary unreliability), and PaS regulates their perturbation intensity along training to realize a non-redundant progress-guided learning trajectory. Although both ScD and BcD adopt structured perturbations, they address different uncertainty sources and are therefore complementary rather than redundant: ScD mainly targets semantic incompleteness inside objects by perturbing target regions while preserving sparse anchors, encouraging context-based semantic and morphological reasoning; in contrast, BcD targets boundary unreliability by selectively perturbing ambiguous edges, reducing premature overfitting to noisy boundary supervision. PaS further differentiates their roles by scheduling their influence over training. Eventually, the perturbed images processed by ScD, BcD, and the progress-aware scheduler are merged to form the conditioning input of the condition encoder. Denoting the ScD and BcD perturbed images by $\tilde{\bm{I}}_s$ and $\tilde{\bm{I}}_b$,  the final conditioning image is
\[
\tilde{\bm{I}} \;=\; \tilde{\bm{I}}_s \,\oplus\, \tilde{\bm{I}}_b,
\]
where $\oplus$ denotes merging the two noises by preserving their union. We will introduce the semantic-concentrated diffusion (ScD), boundary-centralized diffusion  (BcD), and progress-aware scheduler (PaS) in detail in the following sections.


\subsection{Semantic-concentrated Diffusion (ScD)}

\begin{figure}
\centering
\includegraphics[width=0.5\linewidth]{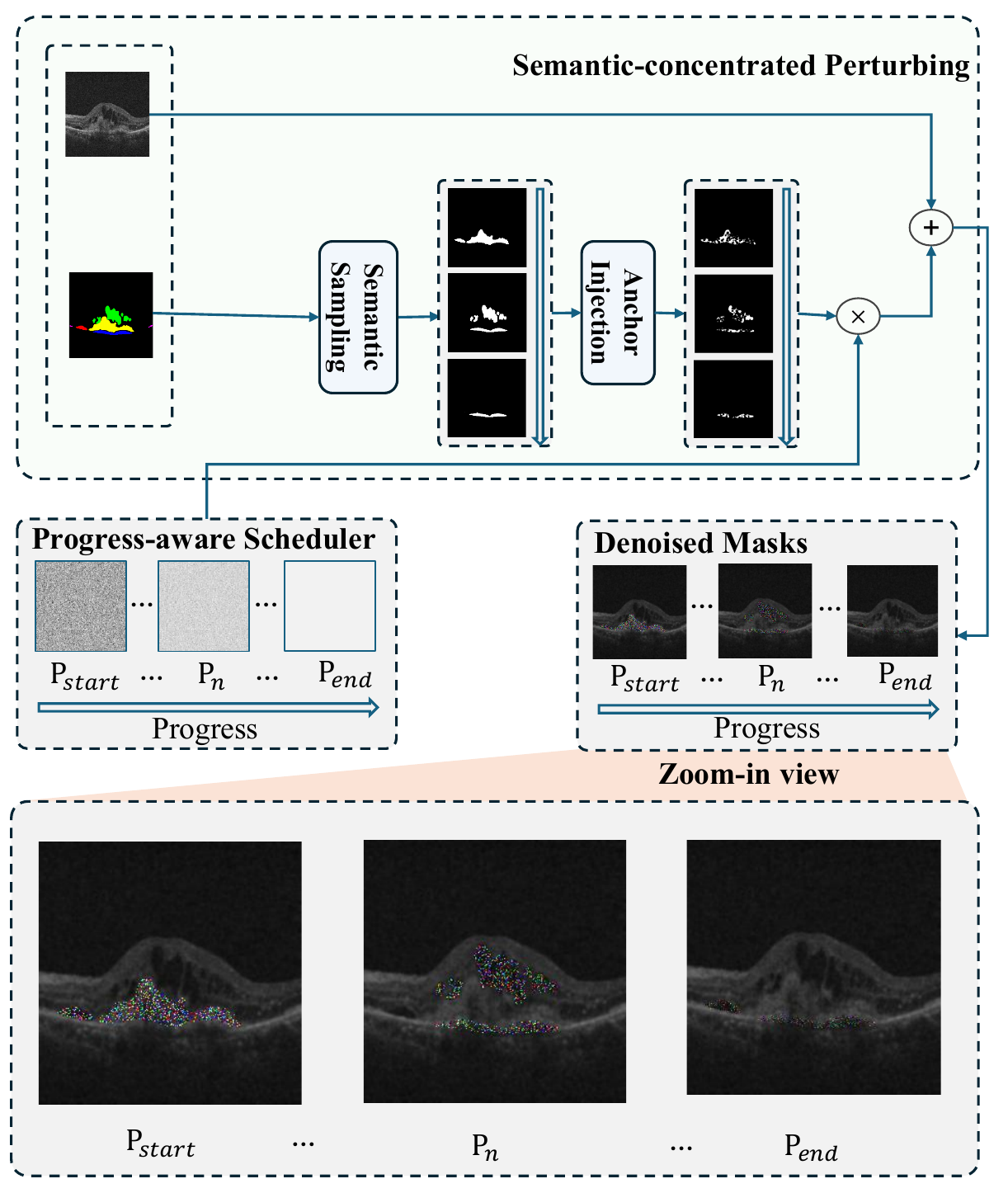}
\caption{Illustration of the semantic-concentrated diffusion (ScD), which perturbs pixels within a specific medical target with progress-aware noise but preserves small unaltered areas as semantic anchors, encouraging structure-guided reasoning and context-based reconstruction.}
\label{component1}
\end{figure}

In our framework, we introduce a semantic-concentrated diffusion (ScD) strategy to enhance the SPAD's ability to capture semantic structures. As shown in Fig.~\ref{component1}, ScD is operationally realized by the semantic-concentrated perturbing (ScP) module and coordinated by the progress-aware scheduler (PaS). The PaS controls the strength of the injected noise over training and gradually reduces it along the training progress ($P$) from the beginning of training ($P_{\text{start}}$) to the end ($P_{\text{end}}$). This strategy first selects target regions from the input image $\bm{I}$ via semantic sampling, and then applies localized noise within these regions while preserving a subset of pixels through anchor injection. The injected anchors serve as stable semantic cues, whereas the perturbed pixels compel the model to exploit surrounding context to infer the missing or corrupted areas.

Assume that there are $N$ medical targets in the input image $\bm{I}$ needing segmentation, in which the $i$-th target is denoted as $C_i$. At each training epoch $p$, a subset of target indices $\mathcal{C}_t \subseteq \{C_1, \dots, C_N\}$ is randomly sampled to be perturbed. The subset size decreases progressively during the training process according to the equation as follows,

\begin{equation}
|\mathcal{C}_t| = \left\lfloor \frac{m}{1 + \gamma \cdot p} \right\rfloor, \qquad (m<N)
\end{equation}
where $m$ denotes the maximum number of perturbed targets at the beginning of training and $\gamma$ controls the decay speed. As a result, more targets are perturbed in the early training stages, while more complete semantic information is provided in later stages. Since the selected targets in medical images differ from batch to batch, the diverse perturbation patterns across samples and batches also prevent the diffusion model from overfitting to specific regions in an image.

For each selected target, we derive a binary mask from the ground-truth label and randomly retain a fixed proportion of pixels (e.g., 30\%) as anchors. While the remaining pixels define the \emph{perturbation mask} $\bm{M_s}$ (1 on perturbed pixels, 0 elsewhere, same size as $\bm{I}$). Gaussian noise is then added \emph{only within} $\bm{M_s}$ to produce the perturbed conditioning image $\tilde{\bm{I}}$:
\begin{equation}
\tilde{\bm{I_s}}=\bm{I}+\bm{M_s}\odot\epsilon,\qquad \epsilon\sim\mathcal{N}(\mathbf{0},\sigma_p^2)
\end{equation}
where $\odot$ denotes element-wise multiplication, $\sigma_p$ is the perturbation intensity at the $p$-th training epoch (progress), which is progress-aware and is modulated by the progress-aware scheduler (PaS). Finally, the perturbed image $\tilde{\bm{I_s}}$ is encoded together with anchors and unaltered regions. As a result, during forward propagation, the model must reconstruct the perturbed regions under the guidance of the sparse semantic cues provided by the anchors. In summary, this training process guides inferring corrupted areas from the surrounding semantic context and improves the inter-target structural reasoning and anatomical rationality.

\subsection{boundary-centralized diffusion (BcD)}
\begin{figure}[t]
\centering
\includegraphics[width=0.5\linewidth]{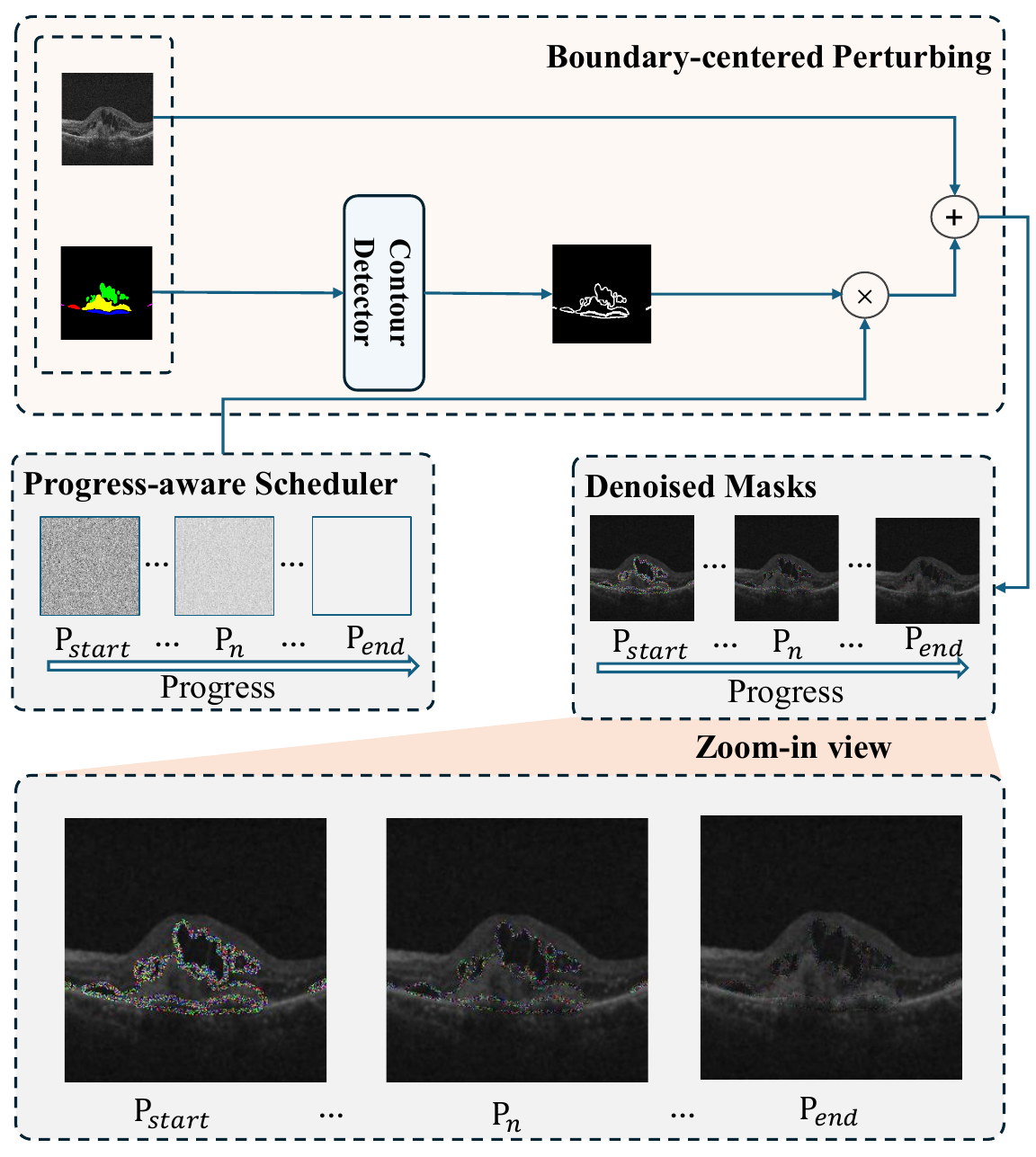}
\caption{Illustration of the boundary-centralized diffusion (BcD). It first extracts the target boundary with a contour detector (such as the Canny operator) from the ground-truth mask label, and then blurs the unreliable and ambiguous boundary with Gaussian noise to lower the reliance on uncertain boundaries during early
target understanding stages.}
\label{component2}
\end{figure}

boundary-centralized diffusion (BcD) is designed to selectively perturb the boundary regions of segmentation labels during training, thereby preventing the model from relying too heavily on ambiguous or uncertain edges. In detail, the BcD is implemented by a boundary-centered perturbing (BcP) module and coordinated by the progress-aware scheduler (PaS). As illustrated in Fig.~\ref{component2}, the strategy begins by constructing a boundary mask from the ground-truth label $x_0$. To achieve this, we first apply a contour detector (such as the Canny operator) with relatively high thresholds so that the extracted results focus on sharp and well-defined contours. The higher thresholds help suppress weak responses caused by noise or annotation inconsistencies, yielding a boundary map that better approximates the regions where boundary uncertainty is most likely to occur.
 
Once the boundary mask is obtained, the Gaussian noise is injected at the corresponding pixels in the input image $\bm{I}$ around the detected boundary, which is similar to the formulation in the ScD strategy. Concretely, letting $\bm{M_b}$ denote the binary boundary mask (where pixels fall on boundary pixels are 1, elsewhere are 0), we perturb boundary locations only, i.e.
\begin{equation}
\tilde{\bm{I_b}}=\bm{I}+\bm{M_b}\odot\epsilon,\qquad \epsilon\sim\mathcal{N}(\mathbf{0},\sigma_p^2)
\end{equation}
Unlike the above ScD, which perturbs the interior of target regions, the BcD produces narrow and elongated perturbations centered along detected contours (see Fig.~\ref{component2}). This makes the corruption pattern sparse and well localized, directly targeting the transition zones between different classes. The perturbed input $\tilde{\bm{I}}$ thus retains the integrity of interior structures while presenting degraded edge information, ensuring that the network processes boundaries under controlled uncertainty. Eventually, the BcD makes the input image remain globally coherent but exhibits band-like corrupted areas around its boundaries. This kind of boundary-centered corruption prevents the diffusion model from memorizing precise edge pixels and encourages it to infer contours from the structural surrounding context.

\subsection{Progress-aware Scheduler (PaS)}
The progress-aware scheduler (PaS) controls the whole learning trajectory (see Fig.~\ref{component1} and Fig.~\ref{component2}) by progressively reducing the overall noise intensity. In other words, the ``progress'' denotes the stage-wise learning progression from coarse structural representation learning to fine boundary refinement. The PaS explicitly encodes this progression by scheduling the perturbation intensity over training epochs, providing a succinct and stable way to coordinate ScD/BcD with the intended coarse-to-fine learning trajectory. During early training stages, the model is exposed to corrupted inputs with high noise intensity, which forces it to focus on learning coarse and stable structures. This allows the model to develop a strong understanding of the overall morphology and semantics and avoid being distracted by fine details at the beginning of training. As the training course continues, the noise intensity gradually decreases, enabling the model to gradually shift to fine and unreliable boundary adjustment during later contour carving stages.

Formally, the noise intensity $\sigma_p$ in Fig.~\ref{component1} and Fig.~\ref{component2}  is controlled by the following decay function epoch by epoch:
\begin{equation}
\sigma_p = \frac{\sigma_{\text{max}}}{1 + \beta \cdot p}
\end{equation}
where $\sigma_{\text{max}}$ is the initial noise intensity, and $\beta$ controls the decreasing rate of perturbing noise. We adopt this inverse-form decay because it could provide a simple, monotonic, and numerically stable attenuation with a controllable long-tail, which smoothly matches the intended progress-guided transition from structure learning to boundary refinement, while remaining the integration simplicity without introducing additional schedule-specific constraints. This decay function ensures that in the early stages of training, the model is exposed to higher levels of noise intensity, which helps it focus on coarse morphological and semantic structures in medical images. As the noise diminishes over time, the model begins to carve the more subtle boundary details, improving segmentation accuracy on challenging cases where fine boundaries of medical targets (such as tumors and lesions) are usually ambiguous and noisy. The same noise intensity is applied to every image batch in a training epoch, ensuring every image undergoes a similar perturbation process throughout a full training epoch.

\begin{table*}[]

\centering
\setlength{\tabcolsep}{2pt}  
\scriptsize
\caption{Comparing with the state-of-the-art methods in medical image segmentation on the AMD-SD dataset}
\begin{tabular}{c|cccccccccccc}
\hline\hline

\multirow{2}{*}{Methods}
  & \multicolumn{6}{c|}{\textbf{mIoU (\%) ↑}}
  & \multicolumn{6}{c}{\textbf{mDice (\%) ↑}} \\ \cline{2-13}
  & SRF & IRF & PED & SHRM & IS/OS & \multicolumn{1}{c|}{Mean}
  & SRF & IRF & PED & SHRM & IS/OS & Mean \\ \hline
U-Net \cite{unet}
  & 64.81 & 64.82 & 66.14 & 56.37 & 72.06 & \multicolumn{1}{c|}{64.48}
  & 78.65 & 78.66 & 79.62 & 72.10 & 83.76 & 78.40 \\ \hline
Swin-Unet \cite{cao2022swin}
  & 69.92 & 58.28 & 64.20 & \textbf{62.75} & 75.65 & \multicolumn{1}{c|}{65.93}
  & 82.30  & 73.64 & 78.20  & \textbf{77.11} & 86.14 & 79.47 \\ \hline
TransUnet \cite{chen2021transunet}
  & 69.49 & 59.81 & 65.19 & 60.29 & 73.39 & \multicolumn{1}{c|}{65.49}
  & 82.00    & 74.85 & 78.93 & 75.23 & 84.65 & 79.15 \\ \hline
MSA \cite{wu2025medical}
  & 55.07 & 45.23 & 54.94 & 55.93 & 49.58 & \multicolumn{1}{c|}{52.03}
  & 71.03 & 62.29 & 70.92 & 71.74 & 66.29 & 68.45 \\ \hline
D-LKA Net \cite{azad2024beyond}
  & 65.59 & 55.32 & 55.30 & 57.29 & 73.42 & \multicolumn{1}{c|}{61.08}
  & 79.22 & 71.23 & 71.22 & 72.85 & 84.67 & 75.84 \\ \hline
DSC-AMP \cite{hu2025dynamic}
  & 71.69 & 66.06 & 74.09 & 60.13 & 73.99 & \multicolumn{1}{c|}{69.02}
  & 83.51 & 79.56 & 85.12 & 75.10  & 85.05 & 81.67 \\ \hline
CCDM \cite{zbinden2023stochastic}
  & 72.96 & 66.98 & 74.70  & 57.75 & 74.54 & \multicolumn{1}{c|}{69.39}
  & 84.37 & 80.23 & 85.52 & 73.22 & 85.41 & 81.93 \\ \hline
\textbf{SPAD (ours)}
  & \textbf{74.24} & \textbf{69.79} & \textbf{76.79} & 59.99 & \textbf{76.72} & \multicolumn{1}{c|}{\textbf{71.51}}
  & \textbf{85.22} & \textbf{82.21} & \textbf{86.87} & 74.99 & \textbf{86.83} & \textbf{83.39} \\ \hline\hline
\end{tabular}
\label{table1}
\end{table*}

\section{Experiments}

\subsection{Datasets}

\subsubsection{AMD‑SD Dataset}
We evaluate our method on the AMD‑SD dataset \cite{hu2024amd}, a high-quality OCT dataset for wet age-related macular degeneration lesion segmentation. It comprises 3,049 B-scan images from 138 patients (156 eyes), each annotated for five lesion-related structures: subretinal fluid (SRF), intraretinal fluid (IRF), ellipsoid‑zone continuity (IS/OS), subretinal hyperreflective material (SHRM), and pigment epithelial detachment (PED). Following the protocols in \cite{hu2024amd}, the data are randomly split into 80\% for training and 20\% for testing at the patient level. The images undergo a multi-stage annotation process involving junior ophthalmologists and senior experts, achieving high inter‑annotator agreement (Dice 0.902–0.979). 

\subsubsection{CXRS Dataset}
The Chest X-ray Segmentation (CXRS) dataset \cite{hu2025dynamic} contains 1,254 high-resolution chest X-ray images (most exceeding \(2\,\mathrm{K}\times2\,\mathrm{K}\)) with pixel-wise annotations for 31 anatomical structures, including 24 ribs, 2 clavicles, 2 scapulae, 2 lungs, and 1 mediastinum. The dataset is divided into training (879), validation (125), and test (250) sets in a 7:1:2 ratio, with each subset maintaining a 3:2 ratio between normal and abnormal (lesion-present) cases.



\subsection{Implementation Details}

We utilize the Adam optimizer~\cite{kingma2014adam} with an initial learning rate
of $4\times10^{-4}$, a weight decay of $1\times10^{-5}$, and a batch size of 8.
The model is trained for 200 epochs on the AMD-SD dataset and 100 epochs on the
in-house CXRS dataset. All input images are normalized to the range $[0,1]$ and
resized to fixed resolutions ($256\times256$ for AMD-SD and $512\times512$ for
CXRS). As described in Section~III, the ScD and BcD are both guided by the PaS,
which adopts a Gaussian noise schedule defined as
$\sigma_p = \sigma_{\mathrm{max}}/(1 + \beta p)$ with
$\sigma_{\mathrm{max}}=0.5$ and $\beta=0.2$. For ScD, the perturbation starts from
3 targets ($m=3$) and gradually decays with $\gamma=0.1$, while 30\% of each
target mask is randomly retained as anchors. 

The whole framework is optimized with a standard conditional categorical diffusion loss \cite{austin2021structured,zbinden2023stochastic}. Specifically, during the intermediate diffusion steps, the loss function consists of a pixel-wise Kullback-Leibler divergence between the predicted reverse transition and true posterior distribution. At the final step, the loss function reduces to a cross-entropy loss based on the ground-truth segmentation. This approach ensures effective training of the diffusion model, allowing it to progressively denoise the corrupted segmentation maps and recover the clean label distributions.

\begin{table}[tbp]
\centering
\caption{Comparing with the state-of-the-art methods in medical image segmentation on the CXRS dataset}
\begin{tabular}{l|cc}
\hline\hline
\multicolumn{1}{l|}{\textbf{Methods}} & \textbf{mIoU (\%) ↑} & \multicolumn{1}{l}{\textbf{mDice (\%) ↑}} \\ \hline
U-Net \cite{unet}                                & 66.85              & 77.35                 \\
Swin-Unet \cite{cao2022swin}                             & 66.37              & 77.12                                \\
TransUNet \cite{chen2021transunet}                             & 67.12              & 77.53                   \\

MSA \cite{wu2025medical}        & 61.03              & 73.27              \\
D-LKA Net \cite{azad2024beyond}                            &  69.53       &  79.54       \\ 
DSC-AMP \cite{hu2025dynamic}                              & 70.34         & 82.58     \\
CCDM  \cite{zbinden2023stochastic}                          & 69.98     & 82.33    \\ \hline
\textbf{SPAD (ours)}                                & \textbf{71.55}     & \textbf{83.42}    \\ \hline\hline
\end{tabular}
    \label{table2}
\end{table}

\subsection{Comparison with State-of-the-art Methods}
To validate the effectiveness of the proposed Structure and Progress Aware Diffusion model (SPAD), we conduct experiments on two medical image segmentation benchmarks, including the AMD-SD dataset and the CXRS dataset. For fairness, all baseline methods are trained and evaluated under the same settings described in the Implementation Details. We report both mean IoU (mIoU) and mean Dice (mDice), with results averaged over multiple runs. The baselines include conventional architectures (U-Net~\cite{unet}, TransUnet~\cite{chen2021transunet}, Swin-Unet~\cite{cao2022swin}), foundation model-based approaches (MSA~\cite{wu2025medical}), structure-aware networks (D-LKA Net~\cite{azad2024beyond}, DSC-AMP~\cite{hu2025dynamic}), and the diffusion-based CCDM~\cite{zbinden2023stochastic}.  

\subsubsection{Experimental Comparison on the AMD-SD Dataset}
The segmentation results on AMD-SD dataset are shown in Table~\ref{table1}. The proposed SPAD achieves the best overall performance, with 71.51\% mIoU and 83.39\% mDice. Compared with the second-best method CCDM~\cite{zbinden2023stochastic}, this corresponds to improvements of +2.12\% and +1.46\%.

Conventional encoder–decoder architectures such as U-Net~\cite{unet}, TransUnet~\cite{chen2021transunet}, and Swin-Unet~\cite{cao2022swin} optimize all pixels uniformly throughout training, without mechanisms to adaptively emphasize
uncertain regions. This uniform treatment often leads to failures on irregular morphologies or blurred contours. By contrast, SPAD employs a progressive learning paradigm that explicitly separates structural representation learning from boundary refinement and schedules their optimization over time, allowing the model to adaptively focus on stable structures first and refine uncertain areas later. As shown in Table~\ref{table1}, while SPAD consistently achieves superior overall performance, we observe a slight performance drop on the SHRM category compared to other categories. The SHRM category often appears as small, fragmented regions with complex local patterns. It makes segmenting this category more challenging since accurate delineation of small and fragmented regions heavily depends on fine-grained local details and precise boundary adjustment. 
Our SPAD framework is designed to improve segmentation stability and structural consistency across categories, particularly in scenarios with diffuse or noisy boundaries, resulting in consistent improvements for most categories, including SRF, IRF, and PED. However, for categories that require a higher focus on fine-grained local details and precise boundary delineation (such as SHRM), the slightly smoother results generated from SPAD are not optimal. This mild performance trade-off for SHRM highlights the robustness of SPAD in dealing with structural uncertainty, but also suggests additional reform for handling small, fragmented targets. We plan to explore this issue in future work by investigating adaptive strategies for different category types.

Foundation model-based approaches such as MSA~\cite{wu2025medical} leverage large-scale pretraining, which provides strong general features but lacks task-specific adaptation. As a result, their performance drops in specialized datasets like AMD-SD, where fine-grained lesion delineation is crucial. SPAD, on the other hand, is designed as a task-oriented framework that incorporates structural priors and boundary refinement into the learning process itself, ensuring both adaptability and precision in domain-specific scenarios. Structure-aware models such as D-LKA Net~\cite{azad2024beyond} and DSC-AMP~\cite{hu2025dynamic} attempt to encode morphological priors through enlarged receptive fields or dynamic convolutional operators. However, they optimize structural and boundary cues in a fully coupled manner, often causing conflicts between global semantics and fine details. SPAD avoids this by decoupling structural learning and boundary refinement into ScD and BcD,
respectively, and coordinating them with the progress-aware scheduler to ensure a stable coarse-to-fine progression. Finally, CCDM~\cite{zbinden2023stochastic}, though sharing the same diffusion backbone as ours, mainly focuses on conditional sampling of label distributions and lacks explicit structural priors or progressive supervision. SPAD extends this line by embedding semantic anchors, introducing boundary-aware perturbations, and controlling the training trajectory with the progress-aware scheduler, which collectively leads to stronger robustness and higher accuracy under ambiguous or missing-class conditions.

\subsubsection{Experimental Comparison on the CXRS Dataset}

Table~\ref{table2} reports the results on the CXRS dataset. SPAD achieves the best performance with 71.55\% mIoU and 83.42\% mDice, surpassing the second-best method CCDM~\cite{zbinden2023stochastic} by +1.57\% and +1.09\%. Respectively, conventional architectures such as U-Net~\cite{unet}, TransUnet~\cite{chen2021transunet}, and Swin-Unet~\cite{cao2022swin} process all pixels in the same way across training, which limits their ability to adapt to irregular shapes and blurred contours. By contrast, SPAD progressively schedules structural and boundary learning, allowing the model to handle reliable regions first and refine uncertain areas later. Foundation model-based methods like MSA~\cite{wu2025medical} benefit from large-scale pretraining but lack effective adaptation to specialized datasets. SPAD differs in being explicitly designed for task-specific segmentation, embedding both structural priors and boundary refinement into the training process. Structure-aware networks such as D-LKA Net~\cite{azad2024beyond} and DSC-AMP~\cite{hu2025dynamic} exploit morphological priors, but they couple coarse and fine cues together, which may cause optimization conflicts. SPAD avoids this by decoupling the two stages and coordinating them through the progress-aware scheduler, ensuring smoother learning. Finally, CCDM~\cite{zbinden2023stochastic}, though based on the same diffusion backbone, focuses only on label distribution modeling and does not employ progressive supervision. SPAD extends this paradigm with noise-guided structural and boundary perturbations, yielding more accurate and robust results on CXRS.  

\begin{table}[]
\centering
\caption{Ablation results on the AMD-SD dataset}
\begin{tabular}{l|cc}
\hline\hline
\multicolumn{1}{l|}{\textbf{Methods}} & \textbf{mIoU (\%) ↑} & \multicolumn{1}{c}{\textbf{mDice (\%) ↑}} \\ \hline
Baseline \cite{zbinden2023stochastic}
& 69.79 & 82.20 \\
ScD with PaS  & 70.93      & 83.00      \\
BcD with PaS  & 71.04      & 83.06      \\
ScD+BcD without PaS   & 44.07 & 61.17 \\
ScD+BcD with PaS (\textbf{ours}) & 71.51 & 83.39 \\
\hline\hline
\end{tabular}
\label{ablation}
\end{table}

\begin{figure}
    \centering
    \includegraphics[width=1\linewidth]{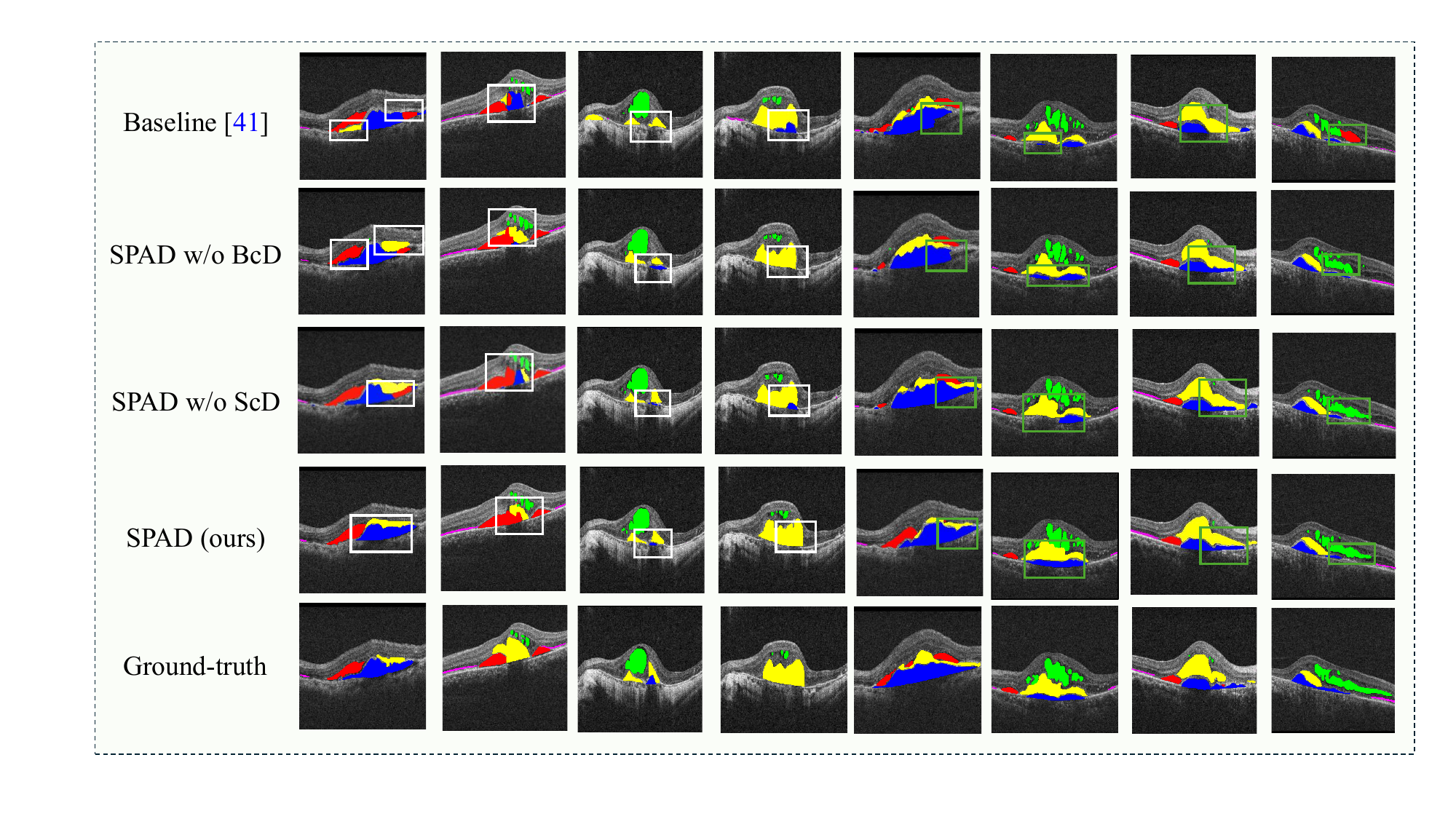}
    \caption{Qualitative visualization of segmentation results on the AMD-SD dataset. From bottom to top: ground truth (GT), segmentation mask predicted from our SPAD, the SPAD without ScD (SPAD w/o ScD), the SPAD without BcD (SPAD w/o BcD), and the diffusion baseline without both ScD and BcD (Baseline). (1) The white boxes in the left four columns indicate representative regions where the baseline model fails to correctly localize target structures or produces incorrect predictions. Compared with the baseline, the methods incorporating ScD exhibit improved structural localization in these regions. (2) The green boxes in the right four columns indicate boundary-sensitive regions where the baseline produces imprecise or irregular contours. Compared with the baseline, the methods incorporating BcD yield more accurate boundary delineation. (3) When both ScD and BcD are incorporated, the predictions combine the advantages of both components and achieve the best.}
    \label{Visualization}
\end{figure}

\subsection{Ablation Studies}
To better understand the effectiveness of each proposed component, we perform ablation studies on the AMD-SD dataset, as summarized in Table~\ref{ablation}. Starting from the base segmentation framework CCDM \cite{zbinden2023stochastic}, we evaluate the individual impact of the proposed components: including ScD (semantic-concentrated diffusion), BcD (boundary-centralized diffusion), and the progress-aware scheduler (PaS). When applying only the ScD and PaS, the performance improves the mIoU from 69.79\% to 70.93\% mIoU and from 82.20\% to 83.00\% for mDice, indicating that preserving anchor regions while perturbing semantic classes helps the model better reason over partial occlusion or class ambiguity. When further applying BcD on top of ScD further enhances the segmentation quality, leading to 71.51\% mIoU and 83.39\% mDice. This shows that boundary-aware perturbation brings additional gains beyond class-level perturbation, and the two strategies together form a complementary design that enables the model to capture both global semantic consistency and fine-grained boundary details, resulting in a more robust and well-structured learning process.

In addition, Table~\ref{Efficiency} reports the computational cost of U-Net, the diffusion baseline, and SPAD under the same hardware setting. As observed, diffusion-based models require longer training and inference time than the classical U-Net framework \cite{unet}, which is mainly attributed to the intrinsic iterative denoising and sampling process. Compared with the diffusion baseline, SPAD introduces only lightweight mask-based perturbation and scheduling operations, resulting in nearly identical training and inference time. This indicates that the proposed structure- and progress-aware strategies incur only marginal additional overhead within the diffusion modeling framework.

\begin{table}[]
\centering
\caption{Efficiency comparison}
\begin{tabular}{l|cc}
\hline\hline
Methods  & Training & Testing \\ \hline
UNet \cite{unet}     & 3.9h     & 0.02h   \\
CCDM (diffusion baseline) \cite{zbinden2023stochastic} & 21.5h    & 0.78h   \\
SPAD (ours)     & 21.6h    & 0.78h   \\
\hline\hline
\end{tabular}
\label{Efficiency}
\end{table}

\subsection{Visualization Analysis}

We further provide qualitative comparisons in Fig.~\ref{Visualization} to illustrate the visual impact of the proposed perturbation strategies. From bottom to top, the figure presents ground truth (GT), segmentation mask predicted from our SPAD, the SPAD without ScD (SPAD w/o ScD), the SPAD without BcD (SPAD w/o BcD), and the SPAD without both BcD and ScD (Baseline).

As shown in the first row, the baseline model frequently yields incomplete or fragmented predictions, particularly in regions with blurred boundaries or subtle lesion appearances. The white-boxed regions highlight cases where the baseline fails to correctly localize target structures or produces incorrect predictions. Compared with the baseline, incorporating ScD improves the recovery of coherent semantic regions and enhances structural localization. The green-boxed regions focus on boundary-sensitive areas, where the baseline predictions exhibit imprecise or irregular contours. In these regions, incorporating BcD leads to more accurate boundary delineation. When both perturbation strategies are applied under the same scheduler, the resulting predictions combine the advantages of both components relative to the baseline. The highlighted regions demonstrate that the scheduler-guided perturbation strategy enables the model to capture both global structure and fine-grained details, yielding results that are visually closer to the ground truth.

\begin{figure*}
    \centering
    \includegraphics[width=1\linewidth]{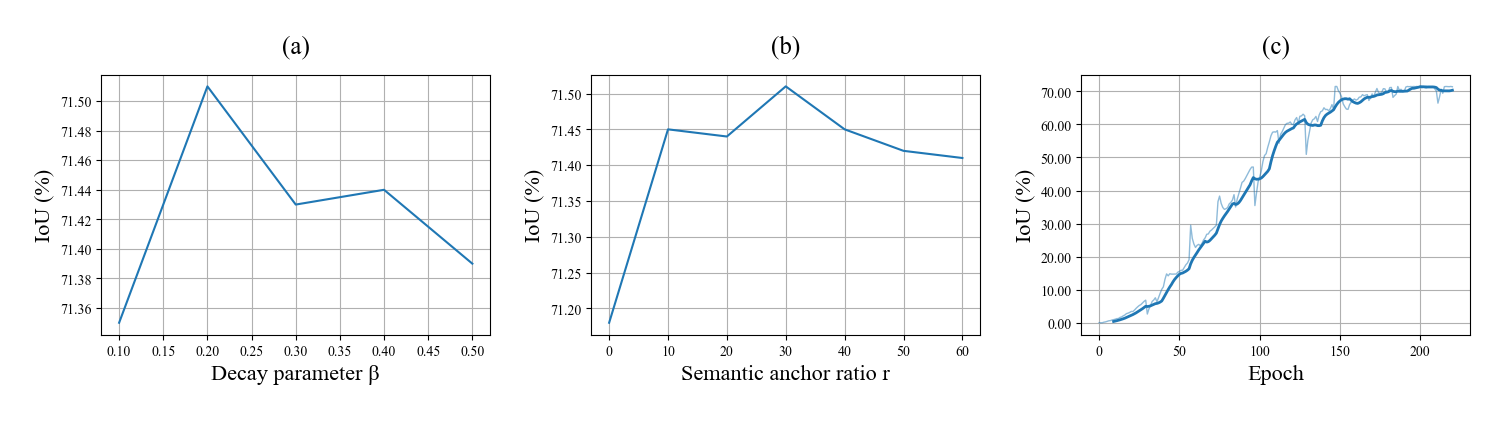}
    \caption{Analysis of the parameter sensitivity and training stability of the proposed SPAD. 
    (a) Sensitivity analysis of the scheduler decay parameter $\beta$ in the progress-aware scheduler (PaS). 
(b) Sensitivity analysis of the semantic anchor ratio $r$ in the semantic-concentrated diffusion (ScD).
(c) Convergence behavior of training the SPAD on the AMD-SD dataset. 
}
    \label{canshufenxi}
\end{figure*}

\subsection{Parameter Sensitivity Analysis}
We conduct sensitivity analysis of the decay parameter $\beta$ in the progress-aware scheduler in Fig.~\ref{canshufenxi}~(a)  . The results show that the segmentation performance varies with different decay rates and reaches a clear peak around $\beta = 0.2$, which is adopted in our main experiments. Both smaller and larger values lead to inferior performance, suggesting that an appropriate decay rate is important for coordinating the transition from early structural learning to later boundary refinement.

We analyze the impact of the semantic anchor ratio of the Semantic-concentrated Diffusion (ScD) in Fig.~\ref{canshufenxi}~(b). The segmentation performance exhibits a non-monotonic trend with respect to the anchor proportion. When the anchor ratio is too small, insufficient semantic references lead to degraded performance, while excessively large ratios weaken the perturbation effect. A moderate anchor ratio (around 30\%) achieves the best overall performance, reflecting a balanced trade-off between semantic preservation and perturbation strength. 

\subsection{Training Stability Analysis}
We analyze the convergence behavior of training SPAD on the AMD-SD dataset in Fig.~\ref{canshufenxi}~(c). The segmentation performance shows a clear upward trend throughout the training process, with moderate fluctuations during the early and intermediate stages, which are common reported in diffusion-based optimization \cite{ho2020denoising,karras2022elucidating}. As training progresses, the curve gradually stabilizes and converges to a steady plateau, indicating the well training stability without any divergence or collapse.

\section{Conclusion}
In this paper, we propose a progress-aware diffusion model for medical image segmentation, built upon a conditional diffusion backbone and coordinated by a progress-aware scheduler (PaS). The framework integrates two complementary perturbation strategies that consist of semantic-concentrated diffusion (ScD) and boundary-centralized diffusion (BcD). The ScD introduces anchor-guided targets perturbation under scheduler control to enhance structural reasoning and improve lesion localization during early coarse Structure stages. The BcD injects boundary-centered perturbation to refine ambiguous contours at later fine boundary stages. Besides, our progress-aware scheduler (PaS) schedules ScD and BcD along training, improving robustness to boundary ambiguity while maintaining structural consistency. Extensive experimental results on the AMD-SD and CXRS datasets demonstrate consistent and significant performance improvement regarding the mIoU and mDice metrics. Despite the consistent improvements achieved by the proposed SPAD, it also faces several limitations remain. On the one hand, 
the fixed-width boundary perturbation design may be less flexible for extremely 
thin or highly elongated anatomical structures, especially during early training stages. Fortunately, our progressive scheduling mechanism partially mitigates this effect over time, 
but more adaptive perturbation strategy could further improve structural flexibility. 
On the other hand, as a naive diffusion-based framework, SPAD inherits the relative computational overhead of iterative denoising. In the future, we will explore adaptive perturbation designs and more efficient diffusion variants to 
enhance both structural generalization and computational efficiency.

\noindent\textbf{Acknowledgments}: This work was supported in part by the National Natural Science Foundation of China (No. 62506003, No. 62376004, No. U24A20342), the Anhui Provincial Natural Science Foundation (No. 2408085QF201), Natural Science Research Major Project of Anhui Provincial Education Department (2025AHGXZK20020), and the Open Project of Anhui Provincial Key Laboratory of Intelligent Detection and Diagnosis for Traffic Infrastructure (No. KY-2025-03).

\bibliographystyle{ieeetr}
\bibliography{ref}

\end{document}